\documentclass[a4paper, 10pt, conference]{ieeeconf}      
\usepackage{FG2023}

\FGfinalcopy 

\usepackage{times}
\usepackage{epsfig}
\usepackage{graphicx}
\usepackage{amsmath}
\usepackage{amssymb}
\usepackage{booktabs}
\usepackage{url}

\usepackage{tikz}
\usepackage{comment}
\usepackage{amsmath,amssymb} 
\usepackage{color}
\usepackage{graphicx}
\usepackage{booktabs}
\usepackage{multirow}
\usepackage{pifont}
\usepackage{float}
\usepackage{times}
\usepackage{epsfig}

\usepackage{amssymb}
\usepackage{pifont}

\definecolor{m_green}{rgb}{0.345 0.502 0.137} 
\definecolor{m_red}{rgb}{0.569 0.071 0.161}

\newcommand{\cmark}{\color{m_green}\ding{51}}%
\newcommand{\xmark}{\color{m_red}\ding{55}}%
\newcommand{\etal}{\textit{et al}.}%

\IEEEoverridecommandlockouts                              
\def\FGPaperID{3} 

\title{\LARGE \bf
Learning from What is Already Out There: \\ Few-shot Sign Language Recognition with Online Dictionaries
}

\begin{document}

\author{\parbox{16cm}{\centering
    {\large Matyáš Boháček$^{1,2}$ and Marek Hrúz$^1$}\\
    {\normalsize
    $^1$ Department of Cybernetics and New Technologies for the Information Society, \\ University of West Bohemia, Pilsen, Czech Republic\\
    $^2$ Gymnasium of Johannes Kepler, Prague, Czech Republic}}
    \thanks{This work has been accepted and scheduled for publication at the \textbf{Face \& Gestures 2023 conference.  }979-8-3503-4544-5/23/\$31.00 \copyright2023 IEEE}
}

\maketitle

\begin{abstract}
Today's sign language recognition models require large training corpora of laboratory-like videos, whose collection involves an extensive workforce and financial resources. As a result, only a handful of such systems are publicly available, not to mention their limited localization capabilities for less-populated sign languages. Utilizing online text-to-video dictionaries, which inherently hold annotated data of various attributes and sign languages, and training models in a few-shot fashion hence poses a promising path for the democratization of this technology. In this work,  we collect and open-source the UWB-SL-Wild few-shot dataset, the first of its kind training resource consisting of dictionary-scraped videos. This dataset represents the actual distribution and characteristics of available online sign language data. We select glosses that directly overlap with the already existing datasets WLASL100 and ASLLVD and share their class mappings to allow for transfer learning experiments. Apart from providing baseline results on a pose-based architecture, we introduce a novel approach to training sign language recognition models in a few-shot scenario, resulting in state-of-the-art results on ASLLVD-Skeleton and ASLLVD-Skeleton-20 datasets with top-1 accuracy of $30.97~\%$ and $95.45~\%$, respectively.
\end{abstract}

\section{Introduction}
\label{sec:intro}

Sign languages (SLs) are natural language systems based on manual articulations and non-manual components, serving as the primary means of communication among d/Deaf communities. While they allow one to convey identical semantics as the written and spoken language, they operate in a distinctively more variable gestural-visual modality. There are currently over $70$ million people worldwide whose native language is one of the approximately $300$ SLs that exist~\cite{WFD}. Nevertheless, no publicly available SL translation system has been introduced so far. This hinders d/Deaf people's ability to use their natural form of communication when working with technology or interacting with people that do not sign. Although the problem of automatic SL Recognition (SLR) has been addressed for many years, it is far from being solved. Modern solutions utilizing deep learning show promise, and neural networks might help tear these barriers down.

There are two prevalent topics related to SLs pursued in the literature - SL Synthesis and SLR. The first one's objective is to translate written language into SL, typically by animating avatars. The second is intended to translate videos of performed signs into the written form of a language. It can be further divided into isolated SLR, which recognizes single sign lemmas out of a known set of glosses, and continuous SLR, translating unconstrained signing utterances. In this paper, we attend to the task of few-shot isolated SLR.

The current methods can be generally divided into two main approaches differing in the means of input representations; the appearance-based and the pose-based. The first prevalent stream of works uses a sequence of RGB images, optionally complemented with the depth channel. These methods reach state-of-the-art results but are more computationally demanding. The second approach performs an intermediate step of first estimating a body pose sequence which is then fed into an ensuing recognition model. These systems tend to be more lightweight and would thus be more suitable for applications on conventional consumer technology, e.g., laptops or mobile phones.

Multiple model training and evaluation datasets have been published over recent years. Generally large-scale in size of glosses and instances, they vary primarily in the originating SL and the manners of data collection. It is essential to consider that, unlike with many tasks in the Natural Language Processing (NLP) domain, no organic sources of potential SL training data (such as the internet and printed media in the case of NLP) yield vast amounts of training instances daily. It hence takes a dedicated, tailored effort to record a SLR dataset. Such an operation is costly and requires specialists from multiple fields at once, making it strenuous and risky to begin with. Accordingly, languages with a smaller user base receive less attention. 

Some of the few resources that contain SL data with built-in annotations are online text-to-video dictionaries. We believe they will be crucial in minimizing barriers in constructing future SLR systems, especially for niche regional contexts. We thus focus on training models using data scraped from such websites. As these services usually contain a few repetitions per sign lemma, such a configuration comprises a few-shot training paradigm. To account for the lack of a diverse, high-repetitive dataset, we utilize SPOTER~\cite{Bohacek_2022_WACV}, a pose-based Transformer~\cite{Vaswani_2017_NEURIPS} architecture for SLR. We hypothesize that it will learn faster since it considers only pre-selected information necessary for such a classification, which is much smaller in dimension than raw RGB video. Appearance-based methods, contrastingly, glutted by the large volume of additional sensory information, need more data to generalize sturdily, as observed by \textit{Boh\'a\v cek}~\etal~\cite{Bohacek_2022_WACV}. We further investigate the ability of models to learn across different datasets and introduce boosting training mechanisms. The main contributions of this work include:

\begin{itemize}
    \item Introducing and open-sourcing UWB-SL-Wild: a new dataset for few-shot SLR obtained from public SL dictionary data, provided with class mappings to already existing SLR datasets;
    \item Proposing Validation Score-Conscious Training procedure which adaptively augments and re-trains for classes that are identified as under-performing during training;
    \item Establishing the state-of-the-art results on the ASLLVD-Skeleton and ASLLVD-Skeleton-20 datasets.
\end{itemize}

\section{Related work}

This section reviews the existing datasets and methods for isolated SLR. As low-instance training has not yet been explored to a greater extent for this task, we consider the overlaps to few-shot or zero-shot gesture and action recognition.

\subsection{Datasets}

Multiple datasets of isolated signs have been published and studied in the literature. We summarize the prominent ones in Table~\ref{tab:datasets}. Purdue RVL-SLLL ASL Database~\cite{Martinez_2002_MMI}, containing $1,834$ videos across $104$ classes within the American Sign Language (ASL), was one of the first to encompass a larger vocabulary. LSA64~\cite{Ronchetti_2016_CACIC} for the Argentinian Sign language is similar in size, as it contains $3,200$ instances from $64$ classes. Later on, substantially larger corpora started to emerge. DEVISIGN~\cite{Chai_2014_CAS}, for instance, provides $24,000$ recordings spanning $2,000$ glosses from the Chinese sign language. Its videos were captured in a laboratory-like environment and were, to the best of our knowledge, the first to provide the depth information along RGB for this task. MS-ASL~\cite{Vaezi_2019_BMVC} brings a similar scale for the ASL, as it contains $25,000$ RGB videos from $1,000$ classes. Lastly, the AUTSL~\cite{Sincan_2022_IEEE} dataset pushed the size and per-class instance ratio even further. It holds $38,366$ RGB-D recordings spanning $226$ classes from the Turkish SL.

While the available datasets span different geographical contexts, most research has centered around ASL. We left out recent datasets, which we consider to capture the most significant traction within the community, from the introductory survey and provide their detailed descriptions below. We later utilize these for experiments and for constructing our new dataset.

\subsubsection{WLASL}

Word-level American Sign Language dataset~\cite{Li_2020_WACV} is a large-scale database of lemmas from the ASL collected from multiple online sources and organizations. The dataset's gloss totals $2,000$ terms with their translations to English. The authors provide training, validation, and test splits. There is an average of over $10$ repetitions in the training set for each class. There are three primary splits of the dataset depending on the number of classes they cover: WLASL100, WLASL300, and WLASL2000. In our experiments, we use the WLASL100 split only.

\begin{table}[ht]
\centering

\caption{Survey of prominent SLR datasets.}
\label{tab:datasets}

\begin{tabular}{l|lllll}
\textbf{Dataset} & \textbf{SL} & \textbf{Gloss} & \textbf{Instances} & \textbf{Format} \\ \hline
\textbf{DEVISIGN \cite{Chai_2014_CAS}}   & CN    & 2,000             & 24,000   & RGB-D       \\
\textbf{LSA64 \cite{Ronchetti_2016_CACIC}}   & AR    & 64             & 3,200   & RGB       \\
\textbf{AUTSL \cite{Sincan_2022_IEEE}}   & TR        & 226            & 38,336  & RGB-D        \\
\textbf{RVL-SLLL \cite{Martinez_2002_MMI}}  & US       & 104           & 1,834     & RGB \\
\textbf{ASLLVD \cite{Neidle_2012_LREC}}  & US       & 2,745           & 9,763     & RGB/Skelet. \\
\textbf{MS-ASL \cite{Vaezi_2019_BMVC}}  & US       & 1,000          & 25,000   &    RGB    \\
\textbf{WLASL \cite{Li_2020_WACV}}   & US       & 2,000          & 21,083   &  RGB

\end{tabular}
\end{table}

\subsubsection{ASLLVD}

American Sign Language Lexicon Video Dataset~\cite{Neidle_2012_LREC} holds $2,745$ classes of unique terms in the ASL. The authors recorded the data in a consistent lab-like environment with a handful of protagonists. The authors have not defined training and testing splits, resulting in an average of nearly $4$ repetitions per gloss in the whole set.

\subsubsection{ASLLVD-Skeleton} \textit{Amorim}~\etal~have later created an abbreviation of the ASLLVD dataset focused on evaluating pose-based methods. They open-sourced pose estimations of all the included videos from OpenPose~\cite{Cao_2019_TPAMI} and proposed fixed training and test splits. The authors also introduced ASLLVD-Skeleton-20, a smaller subset with only 20 classes, enabling computationally resource-lighter and more distinctive ablations studies.

\subsection{Sign language recognition}
The primal works in SLR have leveraged shallow statistical modeling such as Hidden Markov Models~\cite{Starner_1995_MBR, Starner_1998_TPAMI}, which achieved reasonable performance on very small datasets. A big leap has been observed with the advent of deep learning. Convolutional Neural Networks (CNNs) were amidst the first deep architectures employed for this problem~\cite{Camgoz_2020_CVPR, Koller_2016_BMCV, Rao_2018_SPACES, Saunders_2021_IJCV}. These were used to construct unitary representations of the input frames that could be thereafter used for recognition. Later, various Recurrent Neural Networks (RNNs) have been utilized for input encoding as well - namely Long Short-Term Memory Networks (LSTMs)~\cite{Cui_2017_CVPR, Koller_2020_PAMI} or Transformers~\cite{Camgoz_2020_ECCV, Saunders_2021_IJCV}. The usage of different 3D CNNs has also been studied extensively (e.g., with I3D~\cite{Carreira_2017_CVPR, Vaezi_2019_BMVC, Li_2020_WACV}).
\begin{figure*}[ht!]
\begin{center}
\includegraphics[width=1\linewidth]{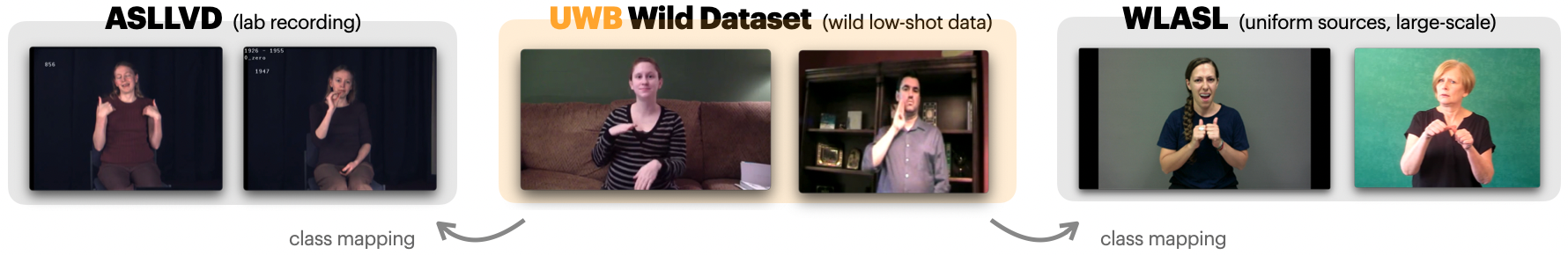}
\end{center}
\caption{Illustrative examples of videos from the used datasets: ASLLVD, WLASL, and our new UWB-SL-Wild. ASLLVD contains videos from a homogeneous lab environment with few repetitions for each class. WLASL consists of videos captured in multiple settings with a larger instance repetition. UWB-SL-Wild, on the other hand, contains videos from an online dictionary with only a handful of examples for each class and both inconsistent signers and recording settings.}
\label{fig:custom_datasets_examples}
\end{figure*}
With the advances in pose estimation, another stream of approaches has emerged, making use of signer pose representations at the input. Unlike the previous methods, these models do not process raw RGB/RGB-D data, but rather pose representations of the estimated body, hand, and face landmarks. \textit{Vázquez-Enríquez}~\etal~\cite{Vazquez_2021_IEEE} have been the first to use a Graph Convolutional Network (GCN) on top of pose sequences, following \textit{Yan}~\etal~\cite{Yan_2018_AAAI} who earlier proposed using GCNs for action recognition. Transformers have been recently employed in this regard, as \textit{Boh\'a\v cek}~\etal~\cite{Bohacek_2022_WACV} introduced Pose-based Transformer for SLR (SPOTER). While the architecture does not surpass the existing appearance-based approaches in general benchmarks, the authors have shown that when trained only on small splits of a training set, SPOTER outperforms even the appearance-based approaches significantly. Lastly, multiple ensemble models combining the raw visual data with the pose estimates~\cite{Jiang_2021_CVPR} have also transpired.

\subsection{Few-shot gesture and action recognition}

Both few-shot gesture and action recognition have not gained extensive traction in literature and are hence not greatly investigated. Most methods have employed metric learning, where the similarity between input videos is learned to classify unfamiliar classes at inference using nearest neighbors. \textit{Bishay}~\etal~\cite{Bishay_2019_BMVC} have proposed the TARN architecture, being the first to incorporate attention mechanism for this task. More recently, Generative Adversarial Networks (GANs) have also been studied in this regard~\cite{Kumar_2019_ICCV}.

\subsection{Few- and Zero-shot SLR}

Zero-shot SLR has been studied by \textit{Bilge}~\etal~\cite{Bilge_2022_TPAMI, Bilge_2019_BMVC}. In both works, the authors propose a pipeline consisting of multiple RNNs and CNNs exploiting the BERT~\cite{Devlin_2019_NAACLHLT} representations of given SL lemmas' textual translations in corresponding primary written language. This has enabled zero-shot SLR to a limited, yet promising extent, supposing the BERT embeddings are available. To the best of our knowledge, the only work addressing few-shot SLR specifically is introduced in~\cite{Wang_2021_AI}. Therein, \textit{Wang}~\etal~leverage a Siamese Network~\cite{Koch_2015_ICML} for feature extraction followed by K-means and a custom matching algorithm.

\section{UWB-SL-Wild}
\label{sec:uwb_sl_wild}

Online SL dictionaries and learning resources are an excellent fit for in-the-wild training data, as they inherently dispose of a gloss annotation. However, since the primary intention with such platforms is not the training of neural networks, only a limited amount of repetitions can be found for each gloss (often $2-3$). To the best of our knowledge, no available benchmark in the literature can simulate such a training paradigm, and we thus decided to create one. We collected a custom dataset called UWB-SL-Wild and are introducing it in this paper.

\begin{figure}[h!]
\begin{center}
\includegraphics[width=1\linewidth]{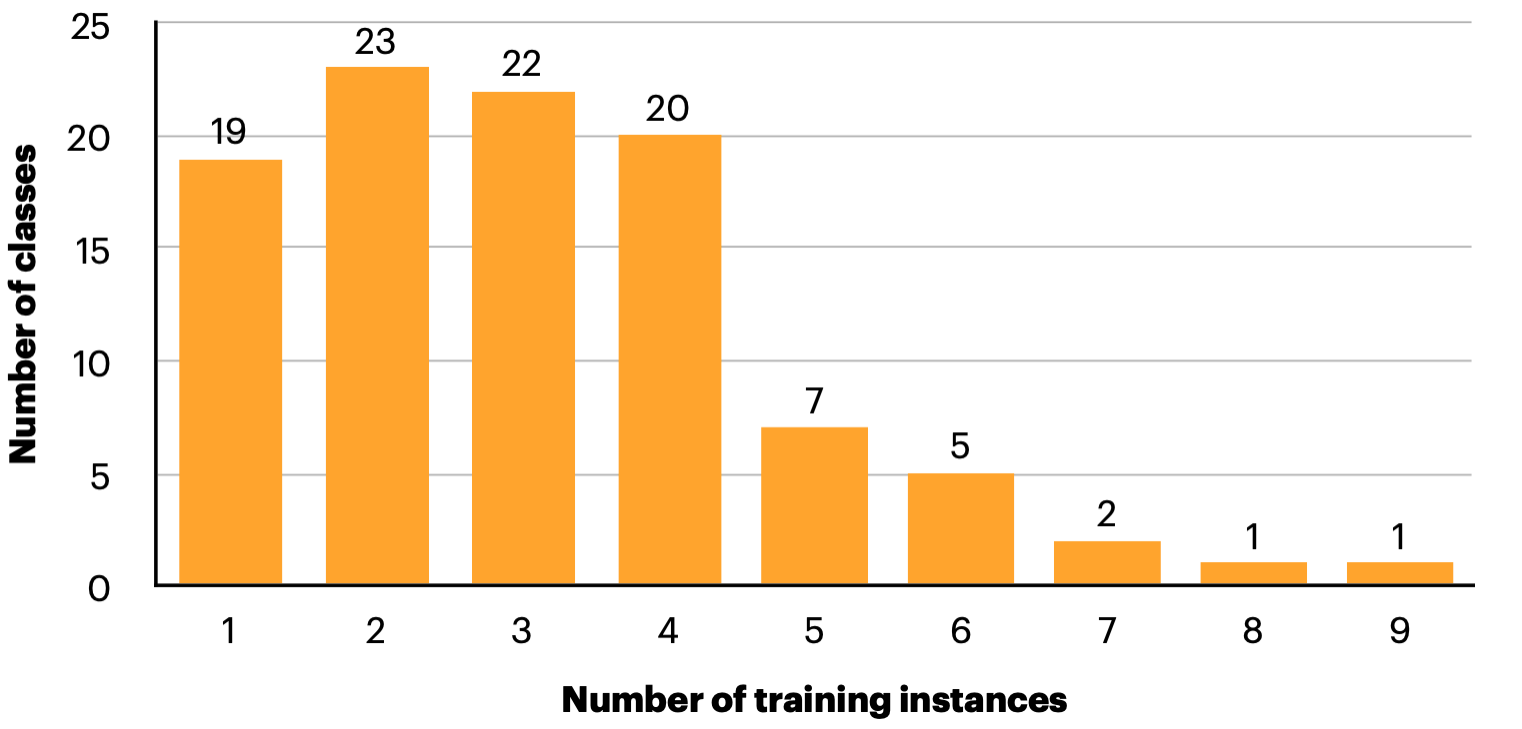}
\end{center}
\caption{Distribution of video repetitions per class in the UWB-SL-Wild dataset. }
\label{fig:class_distrib}
\end{figure}

There are numerous text-to-video dictionaries available on the internet\footnote{As an example, let us mention Spread the Sign (\url{www.spreadthesign.com}), Signing Savvy (\url{www.signingsavvy.com}), Handspeak (\url{www.handspeak.com}), and Sign ASL (\url{www.signasl.org}) websites.}. We decided to use the Sign ASL dictionary as it introduces the largest variability of signer identities and video settings due to gathering videos from multiple providers. The first three websites either contain laboratory-like videos with a single signer (similar to the already existing datasets), have a limited vocabulary, or hold other unsuitable video properties (such as only possessing black-and-white footage).

To allow for transfer learning experiments with the already-existing datasets, we decided that our dataset's vocabulary would be equivalent to that of WLASL100. We then scraped the dataset structure from the Sign ASL portal. This yielded $307$ videos from $100$ classes (corresponding to lemmas in ASL), leaving us with a mean of under $3$ repetitions per class. The total distribution of repetitions per class is depicted in Figure~\ref{fig:class_distrib}. There are $25$ unique signers in the set. Each goes hand-in-hand with a different setting: video quality, distance and angle from the camera, and background. While some stand in front of a wall, many sit casually on a sofa or at a table. We manually annotated the signer identity in each video and are providing this information along with the dataset. The $100$ classes in UWB-SL-Wild represent lemmas of frequent terms in ASL, including ordinary objects (e.g., book, candy, and hat), verbs (e.g., play, enjoy, go), and other adjectives or particles (e.g., thin, who, blue). Given that certain signs in ASL dispose of different variations, it may almost seem as if the signs gathered under a single class were sometimes completely different. Despite distinctively unalike in appearance, they still convey identical or highly similar meanings. This further enlarges the difficulty of learning on this dataset since some glosses' sign variations eventually ended up with only a single instance in the entire set (supposing each of the $2-3$ videos in a given class depicts a different variation). While this is not the case for most classes with just a single variant, a considerable part of the dataset's glosses hold at least two versions. We thus provide manual annotations identifying different variations in each class. We created a mapping schema of classes between UWB-SL-Wild, WLASL100, and ASLLVD datasets\footnote{There were no related videos for $3$ classes of WLASL100 split in \url{SignASL.org}. We thus took the following $3$ classes from the full WLASL to compensate for this.}. This enables future researchers to train on and evaluate using these three datasets. Examples of videos from all three sources can be seen in Figure~\ref{fig:custom_datasets_examples}. We are open-sourcing the UWB-SL-Wild dataset, including the cross-datasets mappings and pose estimates 
of signers in all videos at \url{https://github.com/matyasbohacek/uwb-sl-wild}.

\section{Methods}
This section presents a method that can learn in a few-shot scenario. We build upon SPOTER~\cite{Bohacek_2022_WACV}, as it has shown substantial promise for training on smaller sets of data, fitting our few-shot use case. According to the authors, it should require lower amounts of training data because it is a pose-based method. We review the pipeline's key elements and the changes we have made below. Any unmentioned attributes or configurations were kept identical. We hence refer the reader to the original publication for details.

\textbf{Preprocessing}:
We first estimate the signer's pose in all input video frames. 2-D coordinates of key landmarks are obtained for the upper body ($9$), hands ($2 \times 21$), and face ($70$). 

\textbf{Augmentations and normalization}: 
We follow the augmentation and normalization procedures from~\cite{Bohacek_2022_WACV} to the full extent. 

\subsection{Architecture}

SPOTER is a moderate abbreviation of the Transformer architecture~\cite{Vaswani_2017_NEURIPS}. The input to the network is a sequence of normalized and flattened skeletal representations with a dimension of $242$. Learnable positional encoding is added to the sequence before it is processed further by the standard Encoder module. The input to the Decoder module is a single classification query. It is decoded into corresponding class probabilities by a multi-layer perceptron on top of the Decoder.

\begin{table}[h]
\centering
\begin{centering}

\caption{Performance comparison on ASLLVD-Skeleton dataset}
\label{tab:approach_comparison}

\begin{tabular}{l|ll|ll}
 & \multicolumn{2}{l|}{\textbf{ASLLVD-S}} & \multicolumn{2}{l}{\textbf{ASLLVD-S-20}} \\
\textbf{Model}        & \small top-1       & \small top-5        & \small top-1        & \small top-5         \\ \hline
\textbf{HOF} \cite{Lim_2016_JVCIR}                    & –                     & –                     & 70.0                 & –                      \\
\textbf{BHOF} \cite{Lim_2016_JVCIR}                   & –                     & –                     & 85.0                 & –                      \\ 
\textbf{ST-GCN} \cite{Amorim_2019_ICANN}                & 16.48               & 37.15               & 61.04                & 86.36                \\
\textbf{SPOTER} \cite{Bohacek_2022_WACV}                & 30.77               & 52.05               & 93.18                & 97.72                \\ \hline
\textbf{SPOTER + VSCT}               & \textbf{30.97}                      & \textbf{52.87}                      & \textbf{95.45}                     & \textbf{100.00}  
\end{tabular}
\end{centering}
\end{table}

\subsection{Validation score-conscious training}

In an attempt to adapt the SLR pipeline for the few-shot training environment, we propose the Validation Score-Conscious Training (VSCT). It aims to minimize the classification error on the fly by identifying the bottleneck classes, i.e., the classes that get misclassified the most. VSCT adds the following steps at the end of each epoch of batch gradient descent:

\begin{enumerate}
    \item Validation accuracy is calculated for every class within the set. If a validation split is unavailable, the accuracy is computed on the training split.
    \item The classes are sorted by their performance. A set of classes $W_{vsct}$ is found as a proportion of $\gamma_{vsct} \times c$ worst-performing ones, where $c$ is the total number of classes.
    \item Next, a mini-batch is constructed as a random $\tau_{vsct}$ share of the training set with classes from $W_{vsct}$.
    \item Backpropagation is performed yet again on the above-described mini-batch. However, the parameters of augmentations are drawn from a different distribution. This allows us to target the problematic classes with better-suited representations.
\end{enumerate}

$\gamma_{vsct}$, $\tau_{vsct}$, and all VSCT-specific augmentation parameters are constant hyperparameters of a training run.

\section{Experiments}

In this section, we report our results compared to the already existing methods. We also evaluate our approach on a newly proposed benchmark leveraging the class mappings from UWB-SL-Wild and ASLLVD datasets to WLASL100. 

\begin{table*}[h]
\begin{center}

\caption{Results of the transfer learning experiments where training and evaluation were performed on different datasets}
\label{tab:cross_dataset_experiments}

\begin{tabular}{llll|ll|ll}
 &  &  &  & \multicolumn{2}{l|}{\textbf{ASLLVD → WLASL}} & \multicolumn{2}{l}{\textbf{UWB-SL-Wild → WLASL}} \\

 \textbf{Norm.} & \textbf{Aug.} & \textbf{Bal. sample} & \textbf{VSCT} & test      \hspace{5mm} & val        &  test       \hspace{5mm} & val         \\ \hline
 \xmark  & \xmark & \xmark  & \xmark                        & 10.51 & 5.94 & 8.56 & 7.41                      \\
 \cmark  & \xmark & \xmark & \xmark                         & 19.07 & 16.62 & 14.79 & 16.62                      \\
 \cmark  & \cmark & \xmark & \xmark                         & 19.84 & 15.73 & 15.18 & 16.91                      \\
 \cmark  & \cmark & \cmark & \xmark                         & 20.23 & 15.13 & 16.73 & 14.54                     \\
 \cmark  & \cmark & \xmark & \cmark                         & \textbf{22.96} & 13.95 & \textbf{18.68} & 16.02                   \\
\end{tabular}
\end{center}
\end{table*}

\subsection{Implementation details}
\label{subsec:implementationdetails}

The SPOTER architecture with VSCT has been implemented in PyTorch~\cite{Paszke_2019_NEURIPS}. The model's weights were initialized from a uniform distribution within $\left[0,1\right)$. We trained it for $130$ epochs with an SGD optimizer. The learning rate was set to $0.001$ with no scheduler and both momentum and weight decay set to $0$, following the original implementation. VSCT hyperparameters differ based on the examined dataset. For body pose estimation, we used the HRNet-w48~\cite{Wang_2020_TPAMI} complemented by a Faster R-CNN~\cite{Ren_2015_ANIPS} for person detection within the MMPose library~\cite{Sengupta_2020_Sensors}. We also leveraged the Sweep functionality (hyperparameter search) within the Weights and Biases library~\cite{wandb} to find augmentation and VSCT hyperparameters. We namely employed the Bayesian hyperparameter search method~\footnote{For details on this search method, we refer the reader to the official Weights and Biases documentation available at \url{https://docs.wandb.ai/guides/sweeps/}.} and conducted this procedure for each dataset individually.

\subsection{Quantitative results}

The results on the ASLLVD-Skeleton dataset, along with a comparison to the already available methods, are shown in Table~\ref{tab:approach_comparison}. We establish an overall state-of-the-art on this benchmark by achieving $30.97 \%$ top-1 and $52.87 \%$ top-5 accuracy on the primary dataset. Our method surpasses the pose-based ST-GCN by a significant margin, almost doubling the top-1 performance. When evaluated on the much smaller 20-class subsplit, SPOTER+VSCT achieves $95.45 \%$ top-1 and $100.0 \%$ top-5 accuracy, which exceeds the so far best BHOF by more than absolute $10 \%$. Note that all the models listed in rows 1-5 of Table~\ref{tab:approach_comparison} use appearance-based representations. BHOF, for instance, builds upon a block-based histogram of the incoming videos' optical flow.

The latter of our evaluation settings makes use of the class mappings introduced in Section \ref{sec:uwb_sl_wild}. We trained SPOTER+VSCT on ASLLVD or UWB-SL-Wild dataset but calculated the accuracy on the WLASL100 testing set.
We made the WLASL100 validation split available to the training procedure for the purposes of per-class statistics computation within VSCT. The results are presented in Table~\ref{tab:cross_dataset_experiments}. SPOTER+VSCT achieves a top-1 accuracy of $22.96 \%$ when trained on ASLLVD and $18.68 \%$ when trained using UWB-SL-Wild.

\begin{table}[h]
\centering
\begin{centering}

\caption{Ablation study on ASLLVD-Skeleton dataset}                              
\label{tab:ablation_study}

\begin{tabular}{llll|ll}
&  &  &  & \multicolumn{2}{l}{\textbf{ASLLVD-S}} \\
\small \textbf{Norm.} & \small \textbf{Aug.} & \small \textbf{Bal. sample} & \small \textbf{VSCT} & \small Full & \small 20 cls. \\ \hline
\xmark  & \xmark  & \xmark                  & \xmark             & 5.13  & 47.73                   \\
\cmark  & \xmark  & \xmark                  & \xmark             & 29.18  & 86.36                   \\
\cmark  & \cmark  & \xmark                  & \xmark             & 30.77  & 88.64                   \\
\cmark  & \cmark  & \cmark                  & \xmark             & 30.77  & 90.91                   \\
\cmark  & \cmark  & \xmark                  & \cmark             & \textbf{30.97}  & \textbf{95.45}               
\end{tabular}
\end{centering}
\end{table}

To provide context to these values, let us consider the results of \textit{Boh\'a\v cek} \etal~\cite{Bohacek_2022_WACV} who trained and evaluated SPOTER (without VSCT) on WLASL100. They achieved $63.18 \%$, roughly three times greater accuracy. Their training set averaged $10.5$ repetitions per class, whereas ASLLVD and UWB-SL-Wild have a mean of $3.6$ and $2.9$ per-class instances, respectively. Moreover, UWB-SL-Wild is significantly more variable as opposed to the other two datasets in both unique protagonists and camera settings. While these cross-dataset results are not nearly comparable to the standard methods applied for WLASL100 benchmarking, we believe they attest to the pose-based methods' ability to generalize on characteristically distinct few-shot data.

\subsection{Ablation study}
\label{subsec:ablation_study}

We have conducted an ablation study of the individual contributions of normalization, augmentations, and the VSCT to the above-presented results. We also compare VSCT to the balanced sampling of classes, which counterbalances the disproportion of per-class samples in the training set.
We summarize the ablations on the ASLLVD-Skeleton dataset and its 20-class subset in Table~\ref{tab:ablation_study}. \textit{Norm.}, \textit{Aug.}, and \textit{Bal. sample} refer to using normalization, augmentations, and balanced sampling, respectively, in the given model variant. The baseline models achieved an accuracy of $5.13 \%$ and $46.73 \%$, respectively. We can observe that normalization itself provides the most significant improvement to $29.18 \%$ and $86.36 \%$, while augmentations provide a slight boost on top of that, resulting in an accuracy of $30.77 \%$ and $88.64 \%$.

With all the previous modules fixed, we test the advantages of using either balanced sampling or VSCT. As for the complete dataset, balanced sampling does not provide any performance benefits, whereas VSCT brings a slight improvement resulting in $30.97 \%$ testing accuracy. When examined on the smaller subset, the balanced sampling improves the result by a relative $2.6 \%$ to $90.91 \%$. VSCT, nevertheless, still outperforms it by enhancing the result with a relative $7.7 \%$ to the final $95.45 \%$ testing accuracy.

The outturn of ablations on the cross-dataset training experiments is shown in Table~\ref{tab:cross_dataset_experiments}. For both ASLLVD and UWB-SL-Wild, we conduct the same ablations. The results mimic the tendencies commented on in the previous experiment.
This study suggests that VSCT provides merits to training on such low-shot data, evincing itself more beneficial than a standard balanced sampling of classes.

\section{Conclusion}
We collected and open-sourced a new dataset for SLR with footage from online text-to-video dictionaries. We constructed it with the already-available datasets in mind and created class mappings to WLASL100 and ASLLVD. To reflect the attained problem's few-shot setting, we proposed a novel procedure of training a neural pose-based SLR system called Validation Score-Conscious Training. This procedure analyzes intermediate training results on a validation split and adaptively selects samples from the worst-performing classes to create additional mini-batches for training. We demonstrated VSCT's merits in several experiments of few-shot learning tasks utilizing the SPOTER model, resulting in a state-of-the-art result on the ASLLVD-Skeleton dataset. 

\section*{Acknowledgement}
This work was supported by the Ministry of Education, Youth and Sports of the Czech Republic, Project No. LM2018101 LINDAT/CLARIAH-CZ. Computational resources were supplied by the project ”e- Infrastruktura CZ” (e-INFRA CZ LM2018140).

{\small
\bibliographystyle{ieee}
\bibliography{egbib}
}

\end{document}